\documentclass[conference]{IEEEtran}
\IEEEoverridecommandlockouts
\usepackage{cite}
\usepackage{amsmath,amssymb,amsfonts}
\usepackage{algorithmic}
\usepackage{graphicx}
\usepackage{textcomp}
\usepackage{xcolor}
\usepackage{booktabs}
\def\BibTeX{{\rm B\kern-.05em{\sc i\kern-.025em b}\kern-.08em
    T\kern-.1667em\lower.7ex\hbox{E}\kern-.125emX}}
\makeatletter
\newcommand{\linebreakand}{\end{@IEEEauthorhalign}\hfill\mbox{}\par\mbox{}\hfill\begin{@IEEEauthorhalign}}
\makeatother
\begin{document}
\raggedbottom

\title{The 2026 ACII Dyadic Conversations (DaiKon) \\ Workshop \& Challenge\\
}

\author{
\IEEEauthorblockN{Panagiotis Tzirakis}
\IEEEauthorblockA{\textit{Hume AI} \\
New York, USA \\
panagiotis@hume.ai}
\and
\IEEEauthorblockN{Alice Baird}
\IEEEauthorblockA{\textit{Hume AI} \\
New York, USA \\
alice@hume.ai}
\and
\IEEEauthorblockN{Jeffrey Brooks}
\IEEEauthorblockA{\textit{Hume AI} \\
New York, USA \\
jeff@hume.ai}
\and
\IEEEauthorblockN{Emilia Parada-Cabaleiro}
\IEEEauthorblockA{\textit{University of Music Nuremberg} \\
Nuremberg, Germany \\
\scriptsize emiliaparada.cabaleiro@\\\scriptsize hfm-nuernberg.de}
\and
\IEEEauthorblockN{Lukas Stappen}
\IEEEauthorblockA{\textit{BMW Group} \\
Munich, Germany \\
lukas.stappen@bmw.de}
\linebreakand
\IEEEauthorblockN{Sharath Rao}
\IEEEauthorblockA{\textit{Hume AI} \\
New York, USA \\
sharath@hume.ai}
\and
\IEEEauthorblockN{Theo Lebryk}
\IEEEauthorblockA{\textit{Hume AI} \\
New York, USA \\
theo@hume.ai}
\and
\IEEEauthorblockN{Jakub Piotr C\l{}apa}
\IEEEauthorblockA{\textit{Hume AI} \\
New York, USA \\
jakub@hume.ai}
\and
\IEEEauthorblockN{Jens Madsen}
\IEEEauthorblockA{\textit{Hume AI} \\
New York, USA \\
jens@hume.ai}
}


\maketitle

\begin{abstract}
The 2026 ACII Dyadic Conversations (ACII-DaiKon) Workshop \& Challenge introduces
a benchmark for modeling interpersonal affect and social dynamics in dyadic
conversations. Although conversational affect modeling has advanced rapidly,
most benchmarks remain speaker-centric and underrepresent coupled,
time-evolving processes between partners, including directional influence,
conversational timing coordination, and rapport development. To address this
gap, ACII-DaiKon presents three coordinated sub-challenges built on a shared
dataset: (1) directional interpersonal influence prediction, (2) turn-taking
prediction (next-speaker and time-to-next-speech), and (3) rapport trajectory
prediction across full interactions.

The challenge is built on the Hume-DaiKon dataset, comprising 945 dyadic
conversations (743.4 hours of audiovisual data) collected under naturalistic
conditions across five languages. The benchmark supports multimodal modeling,
temporal reasoning, and cross-context generalization through fixed
train/validation/test splits, standardized metrics, and released baseline
systems. Evaluation uses Concordance Correlation Coefficient (CCC), Pearson
correlation, Macro-F1, and Mean Absolute Error (MAE) depending on the
sub-challenge.

Baseline experiments establish initial reference performance, with best test
results of 0.40 CCC and 0.50 Pearson for influence prediction, 0.66 Macro-F1
and 1.50~s MAE for turn-taking, and 0.68 CCC and 0.70 Pearson for rapport
trajectory modeling. These results indicate that while current methods capture
coarse dyadic patterns, robust modeling of directional dependence and
long-horizon interpersonal dynamics remains challenging. The workshop provides
a shared platform for rigorous comparison and cross-disciplinary discussion on
data validity, evaluation protocols, and culturally aware modeling for dyadic
interaction.

\end{abstract}


\section{Introduction}
Dyadic conversation is a core setting in which affect is expressed, perceived, and regulated between individuals~\cite{zaki2013interpersonal}. In face-to-face interaction, people continuously respond to each other through speech, timing, prosody, facial behavior, and other social signals. These processes are inherently relational: one speaker’s behavior shapes the other’s affective state, influences when turns begin and end~\cite{sacks1974turntaking}, and contributes to the development of rapport across the interaction~\cite{tickledegnen1990rapport}. Despite this, many existing affective computing benchmarks and shared tasks continue to frame prediction at the level of isolated speakers or static conversation summaries. As a result, several central aspects of interpersonal interaction remain underexplored in a common evaluation setting, including directional influence, turn exchange, adaptation over time, and the development of rapport.

This gap is consequential from both scientific and computational perspectives~\cite{zaki2013interpersonal,sacks1974turntaking,tickledegnen1990rapport}. From a modeling standpoint, dyadic interaction requires systems that can represent temporal dependence, inter-speaker coupling, and multimodal context across multiple timescales \cite{baltrusaitis2019multimodal, tzirakis2017end}. From an application standpoint, advances in this area are directly relevant to socially aware conversational AI, human--computer interaction, behavioral analysis, and affective dialogue systems~\cite{picard1997affective, baird2022icml}. A benchmark that jointly evaluates these dimensions within a common experimental setting can therefore enable more systematic comparison across methods and encourage architectures that move beyond speaker-centric prediction.

In this paper, we introduce the 2026 ACII Dyadic Conversations (DaiKon) Workshop \& Challenge, a competition-based workshop focused on modeling affective and social dynamics in dyadic interaction. The challenge comprises three sub-challenges built on a single corpus of naturalistic dyadic conversations, allowing participants to study complementary aspects of interpersonal behavior from a common data source. The first, DaiKon Influence, targets target-speaker affect prediction from dyadic context. The second, DaiKon Turn-Taking, targets next-speaker and next-onset prediction. The third, DaiKon Rapport, targets the prediction of rapport trajectories over the course of a conversation. Together, these tasks are intended to capture local coordination, directional dependence, and longer-horizon interaction quality within a unified evaluation framework.

The challenge is based on the Hume-DaiKon corpus. The current public release contains 945 sessions totaling 743 hours, with train, validation, and test splits of 661, 142, and 142 sessions, respectively. According to the public challenge materials, the train and validation partitions contain conversations in German, English, Spanish, Dutch, and Polish. The benchmark was designed to preserve the multilingual composition of the corpus rather than restrict it to one or two languages, while recognizing residual imbalance across languages and speaker demographics.

The challenge also provides baseline methods for all three tasks and uses held-out test labels for official evaluation. These baselines establish initial reference performance, with the strongest test results reaching 0.40 CCC and 0.50 Pearson for influence prediction, 0.66 Macro-F1 and 1.50~s MAE for turn-taking, and 0.68 CCC and 0.70 Pearson for rapport prediction. By combining fixed data splits, shared evaluation protocols, and released baselines, the workshop is designed to support reproducible comparison and to stimulate discussion about robust, culturally aware modeling of dyadic interaction.

\section{The Competition Dataset}
The Hume-DaiKon corpus is a dataset of naturalistic dyadic interaction collected with a dual-channel conversational platform. The current challenge release contains 945 sessions totaling 743.4 hours, partitioned into 661 training, 142 validation, and 142 test sessions. The benchmark uses multilingual data from German, English, Spanish, Dutch, and Polish, with splits stratified by language and with the test set kept blind for official evaluation.

The data were collected using a conversational platform developed by Hume AI that connects pairs of participants from similar geographic regions. Prior to participation, users complete an audio quality screening to ensure a minimum standard of microphone clarity. They are then asked to respond to a short prompt in their native language (e.g., “How was your weekend?”), and their response is automatically checked using a language model to verify that it is both relevant to the prompt and linguistically fluent. Participants who pass this stage are paired with another user and engage in an open-ended conversation guided by a set of prompts (e.g., “Where did you go on your last vacation?”) designed to elicit a range of expressive and conversational behaviors. Conversations are not time-limited, allowing participants to interact naturally for as long as they choose.

The dataset is designed to support several complementary aspects of dyadic interaction from the same source material, including affect prediction conditioned on both speakers’ history, conversational timing and floor exchange, and the development of rapport over time. The benchmark was designed to preserve the multilingual character of the data rather than restrict it to one or two languages, while still aiming for reasonable balance across language groups in the released partitions.

The participant release includes the combined conversational audio signal together with separate per-speaker video streams, while internal metadata used for preprocessing and label generation are not part of the public release. For the released baselines, raw audio is stored in \texttt{.wav} format and raw video in \texttt{.webm} format. Audio features are extracted with the Whisper-small encoder~\cite{radford2023whisper} from the combined conversation audio, yielding 768-dimensional frame-level representations at approximately 3 FPS, and visual features are extracted with FaceNet~\cite{schroff2015facenet} from face images, yielding 512-dimensional embeddings at 3 FPS together with bounding box coordinates and face-detection confidence scores. Only video frames with face-detection confidence of at least 0.9 are retained. The released feature files are stored in Parquet format, one file per conversation, so that participants can load only the required modalities and temporal segments without decoding all raw media.

\begin{figure}[t]
\centering
\IfFileExists{figures/daikon_dataset_grid.png}{%
    \includegraphics[width=0.9\columnwidth]{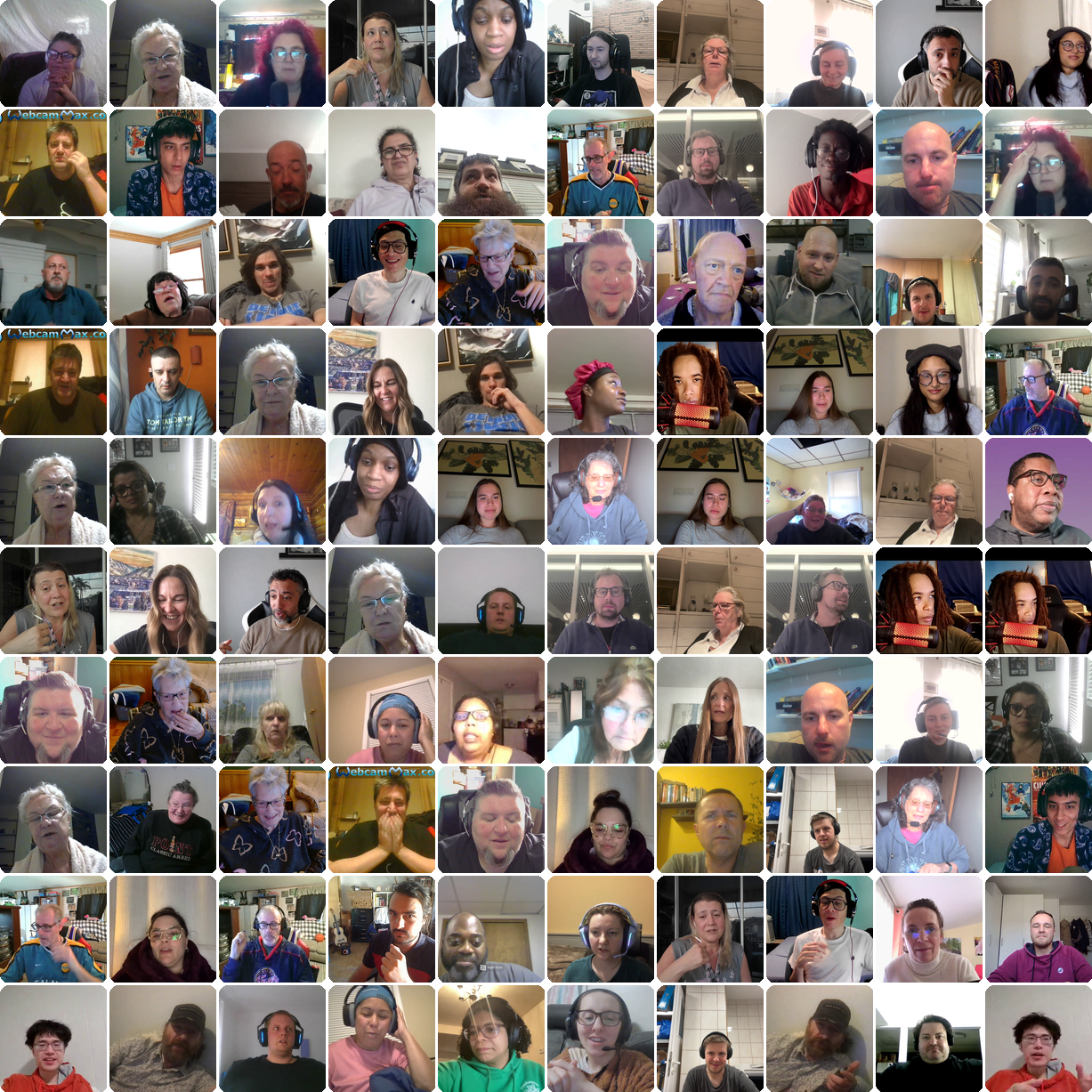}%
}{%
    \fbox{\parbox[c][1.5in][c]{0.9\columnwidth}{\centering Dataset image placeholder\\[2pt]\footnotesize Add \texttt{figures/daikon_dataset_grid.png} to use the challenge-page frame grid.}}%
}
\caption{Example frames from the Hume DaiKon challenge dataset page (Figure~2 on the public challenge website).}
\label{fig:dataset_grid}
\end{figure}

To reduce noise, sessions with missing required files, unavailable voice activity detection, very short duration, or incomplete video were treated as candidates for removal during release preparation. Figure~\ref{fig:dataset_grid} first illustrates example sessions from the public challenge materials, highlighting the multi-camera, face-to-face nature of the corpus and the variability in participants, backgrounds, and recording conditions across conversations.

\begin{figure}[t]
\centering
\IfFileExists{figures/daikon_arousal_valence.png}{%
    \includegraphics[width=0.9\columnwidth]{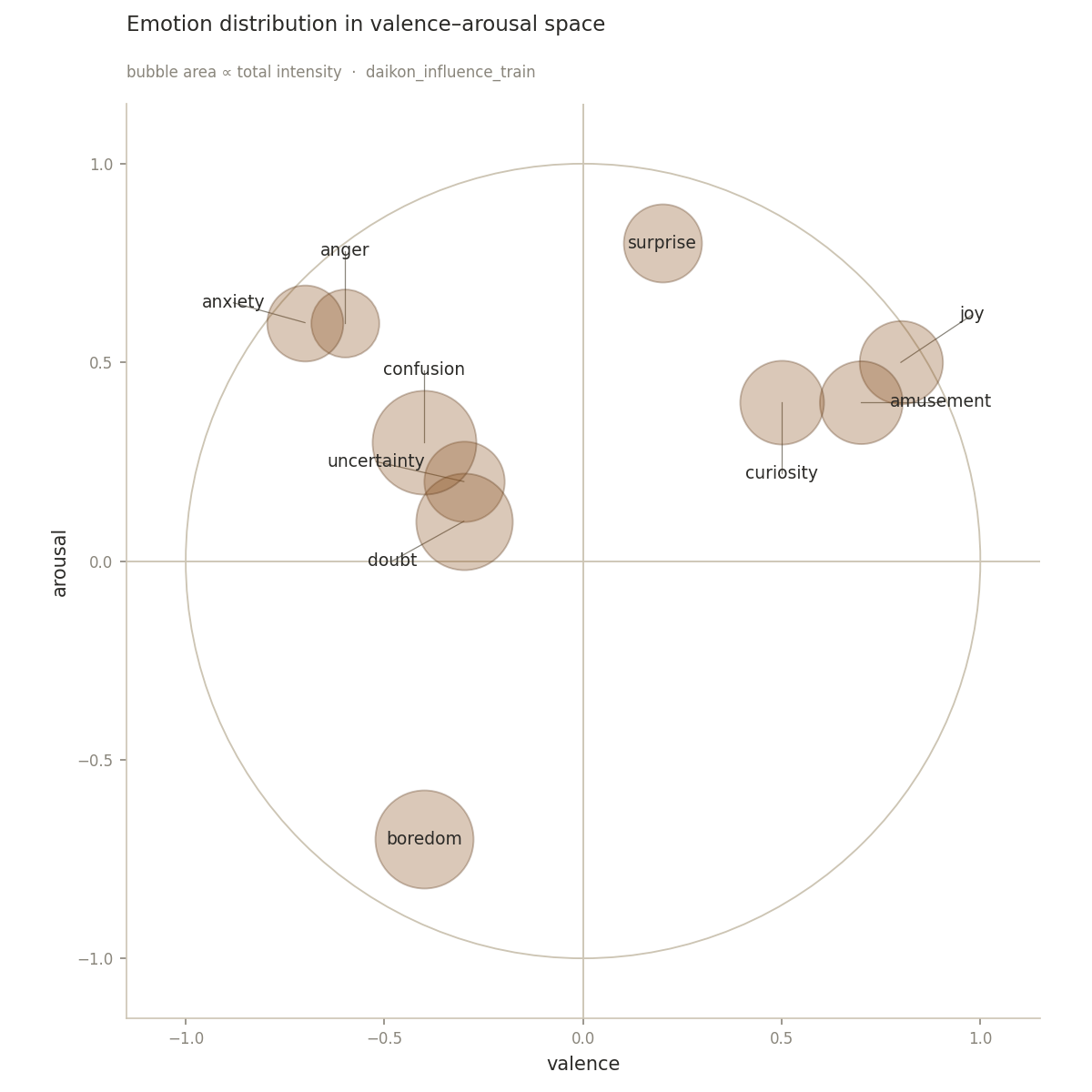}%
}{%
    \fbox{\parbox[c][1.5in][c]{0.9\columnwidth}{\centering Arousal--valence placeholder\\[2pt]\footnotesize Add \texttt{figures/daikon_arousal_valence.png} to display the annotation-space figure.}}%
}
\caption{Arousal--valence visualization from the public Hume DaiKon challenge materials.}
\label{fig:arousal_valence}
\end{figure}

Figure~\ref{fig:arousal_valence} projects the aggregate emotion intensity distribution of the released influence training data onto Russell's circumplex model of affect~\cite{russell1980circumplex}, with bubble area proportional to the summed soft-label intensity of each target emotion across all segments. The ten target dimensions cover three of the four quadrants of the valence--arousal plane, with high-arousal positive states (joy, amusement, surprise, curiosity), high-arousal negative states (anger, anxiety, confusion, uncertainty, doubt), and low-arousal negative states (boredom) all represented. The low-arousal positive quadrant (e.g., calmness, contentment) is intentionally underrepresented, as the selected dimensions were chosen to capture affective states that are informative for interpersonal influence and that are reliably expressed in short conversational segments. This coverage is methodologically important for the sub-challenges: directional influence prediction requires sufficient variance in the target speaker's affective state for inter-speaker coupling to be identifiable, and rapport trajectory modeling relies on supervision that spans both low- and high-engagement interactions. A distribution collapsed around a single point would risk inflating within-sample metrics such as CCC without reflecting genuine generalization across affective conditions.

Finally, Table~\ref{tab:data_splits} summarizes the released partitions, their durations, and participant demographics. The train, validation, and test splits are balanced at the session level for the released benchmark, retain broad geographic diversity, and keep the test set blind for official evaluation.

\begin{table}[t]
\centering
\caption{Hume-DaiKon dataset splits. Test annotations are withheld for competition evaluation. Participant counts are summarized by split, and the language rows report session counts by language.}
\label{tab:data_splits}
\small
\setlength{\tabcolsep}{3.5pt}
\begin{tabular}{lrrrr}
\toprule
\textbf{Statistic} & \textbf{Train} & \textbf{Val} & \textbf{Test} & \textbf{Total} \\
\midrule
\multicolumn{5}{l}{\textit{Overall}} \\[2pt]
Duration (hours) & 504.3 & 118.9 & 120.2 & 743.4 \\
Sessions & 661 & 142 & 142 & 945 \\
Participants & 1322 & 284 & 284 & 1890 \\
\midrule
\multicolumn{5}{l}{\textit{Participants by gender}} \\[2pt]
Female & 530 & 122 & 115 & 767 \\
Male & 779 & 161 & 167 & 1107 \\
Non-binary & 8 & 1 & 1 & 10 \\
Other / unknown & 5 & 0 & 1 & 6 \\
\midrule
\multicolumn{5}{l}{\textit{Sessions by language}} \\[2pt]
German (de) & 46 & 10 & 11 & 67 \\
English (en) & 328 & 75 & 78 & 481 \\
Spanish (es) & 111 & 26 & 21 & 158 \\
Dutch (nl) & 67 & 11 & 13 & 91 \\
Polish (pl) & 107 & 18 & 19 & 144 \\
Other & 2 & 2 & 0 & 4 \\
\bottomrule
\end{tabular}
\end{table}

\subsection{Influence Label Collection}
Influence labels are derived from an internal speech expression model that
produces per-utterance intensity scores across a large vocabulary of emotional
and affective states. For each VAD segment, the model scores both speakers,
enabling directional modeling of interpersonal influence (A$\rightarrow$B and
B$\rightarrow$A). The label vocabulary was chosen by the top-$k$ most highly activated expressions and as best possibly designed to provide balanced
coverage across the valence--arousal space, to reduce biasing the label set toward
high-arousal states. The final ten target dimensions — anger, anxiety,
uncertainty, confusion, doubt, boredom, surprise, curiosity, joy, and
amusement — were selected to span both positive and negative valence and a
range of activation levels, as illustrated in Figure~\ref{fig:arousal_valence}.

\subsection{Turn-Taking Label Collection}
Turn-taking labels are derived from per-speaker VAD applied to the individual
speaker channels. Sessions in which either speaker had empty VAD were excluded entirely, ensuring both participants are active in every retained session. For each
session, a conversational event table was constructed capturing the next
speaker, onset time, and transition type. Three event types are annotated:
\textit{switch} (clean floor transfer with a gap), \textit{overlap}
(simultaneous speech before a transfer), and \textit{interruption} (floor
taken before natural completion). In the English partition alone (278
sessions), events total approximately 257,000 across all types — overlaps
being the most frequent, followed by interruptions and switches — with a mean
of roughly 900 events per session. During release preparation, annotations
were standardized into a consistent speaker-transition schema using the column
names \texttt{from\_speaker} and \texttt{to\_speaker}, \texttt{start\_time},
\texttt{end\_time}, \texttt{switch\_time}, and \texttt{gap\_duration}. The
released benchmark focuses on a cleaner subset of these events to support
reliable classification and regression targets.

\subsection{Rapport Pseudo-label Generation}
\label{sec:rapport-labels}
\label{sec:pseudolabels}
Ground-truth rapport labels are expensive to collect at scale, as they require
trained annotators to provide continuous ratings over full conversation
recordings. To enable training on the full corpus, we adopt a pseudolabeling
approach in which a large language model (LLM) produces per-window rapport
scores from multimodal conversation descriptors.

Each conversation is accompanied by per-speaker diarized VAD segments, each
containing an ASR transcript, precise timestamps, and a vector of 605 acoustic
and emotional tag scores produced by Hume AI's speech expression model~\cite{tzirakis2023largescale}. The
tag vocabulary spans affective states (e.g., \textit{joy}, \textit{boredom}),
prosodic descriptors (e.g., \textit{monotone}, \textit{energetic}), and voice
quality attributes (e.g., \textit{breathy}, \textit{flat}).

Conversations are segmented into overlapping windows of 30 seconds with a
stride of 15 seconds. For each window, we prompt Qwen2.5-72B-Instruct~\cite{qwen2025}
via a vLLM~\cite{kwon2023vllm} server with: (i) the transcripts of the four
preceding windows as conversational context; (ii) the current window
transcript with speaker labels; and (iii) a per-speaker summary of the top-8
rapport-relevant acoustic tags, computed as the duration-weighted mean of
utterance-level tag scores within the window. The model is instructed to score
rapport in the current window only, relative to prior context, on a continuous
$[0,1]$ scale, and to return a JSON object with a score, confidence, and
one-sentence reasoning. Raw scores are smoothed with a Gaussian kernel
($\sigma = 1.5$ windows) to produce trajectories with realistic temporal
dynamics. Of the 1,046 conversations, 804 (76.9\%) have the required
per-speaker metadata and are successfully labeled.

\section{The Competition Tasks}
\label{sec:tasks}
Within this first iteration of ACII-DaiKon, participants are presented with three dyadic sub-challenges built on the Hume-DaiKon dataset, reflecting the growing emphasis on dyadic interaction modeling and multimodal affective computing in recent literature~\cite{javed2024dyadic,song2025affective, stappen2020muse, christ2023muse}.

\subsection{DaiKon Influence}
In the DaiKon Influence sub-challenge, participants are challenged with predicting a target speaker’s affective state from the multimodal dyadic conversational context. In the public challenge formulation, systems predict continuous intensity values for 10 emotion dimensions: anger, anxiety, uncertainty, confusion, doubt, boredom, surprise, curiosity, joy, and amusement, for each labeled speech segment. The evaluation measures are Concordance Correlation Coefficient (CCC) and Pearson correlation, averaged over the target dimensions.

This sub-challenge is designed to move beyond speaker-isolated prediction by encouraging models that capture how one participant’s preceding behavior relates to the affective state of the other participant.

\subsection{DaiKon Turn-Taking}
In the DaiKon Turn-Taking sub-challenge, participants are challenged with predicting both who speaks next and when the next speech onset occurs. The task therefore combines classification and regression components within a single conversational timing benchmark. The official measures are Macro-F1 and accuracy for next-speaker prediction, and Mean Absolute Error (MAE) for time-to-next-speech.

For this task, audio features are particularly relevant because they encode pause structure, speech activity, prosodic continuation, and local rhythmic patterns that often precede turn transitions. Video features can capture gaze, facial orientation, and visible preparatory behavior associated with yielding or taking the floor. The multimodal setting encourages participants to model how these auditory and visual signals jointly support floor-exchange prediction in dyadic interaction.

This sub-challenge targets a central property of dyadic interaction, namely the local coordination of floor exchange.

\subsection{DaiKon Rapport}
In the DaiKon Rapport sub-challenge, participants are challenged with predicting rapport as it evolves over the course of a conversation. Rather than predicting only a single global interaction score, systems output rapport values for labeled windows, enabling trajectory-level modeling of conversational quality over time. The official measures are CCC and Pearson correlation between predicted and reference rapport values.

This task is naturally long-horizon and can benefit from audio descriptors related to speaking style, vocal warmth, intensity, and synchrony, as well as video cues related to facial expressivity, head movement, and visible behavioral alignment. Because rapport develops over time, the task also favors models that aggregate information over longer temporal windows rather than relying only on local frame- or turn-level evidence.

This sub-challenge is intended to encourage models that capture both gradual and abrupt changes in interaction quality as conversations unfold.

Together, these sub-challenges are designed to encourage models that capture interdependence, temporal structure, and robustness across conversational contexts.

\subsection{General Guidelines}
To participate in the ACII-DaiKon competition, participants are required to complete the relevant Hume end-user license agreement and follow the data-access instructions provided on the official competition homepage. In addition, each team must submit a workshop paper describing its methods and results in accordance with the official ACII submission guidelines; the workshop also welcomes related contributions beyond the shared-task entries.

To obtain official test scores, participants submit predictions for the held-out test set to the competition organizers, who evaluate the most recent submission under the official protocol. Teams may participate in any one of the three tasks or compete in all of them. To support reproducibility, participants are also encouraged to upload their code alongside the paper submission, and the organizers may reproduce and further evaluate selected systems.

\section{Baseline Methods}
The public challenge website reports released baseline systems for audio, video, and multimodal modeling across the three tasks. In general, the released baselines are intended as reference implementations rather than highly optimized challenge entries. They provide standardized starting points for data processing, feature extraction, model training, and evaluation under the official splits.

\subsection{Shared Feature Extraction}
All three competition tasks use the same released feature pipeline. Audio input is represented with Whisper-small speech embeddings~\cite{radford2023whisper} extracted from raw \texttt{.wav} recordings at approximately 3 FPS, yielding 768-dimensional representations that capture prosodic and paralinguistic information relevant to affect and interaction dynamics. Video input is represented with 512-dimensional FaceNet embeddings~\cite{schroff2015facenet} extracted at 3 FPS from face images derived from the raw \texttt{.webm} recordings, together with bounding boxes and detection confidence scores. These shared representations support audio-only, video-only, and multimodal baselines, allowing each task to exploit vocal cues, visible behavior, and dyad-level temporal context within a consistent preprocessing framework.

To reduce noise in the released features and labels, sessions were skipped whenever all required files were not present or when voice activity detection was unavailable for one of the speakers. Very short conversations and incomplete rooms, including cases without usable video, were also identified as candidates for removal during release preparation. This preprocessing choice favors cleaner supervision and more reliable alignment across modalities, although it also reduces the number of usable sessions relative to the raw collection.

\subsection{Released Baselines}
For DaiKon Influence, the reported test performance is 0.40 CCC / 0.50 Pearson for audio, 0.19 / 0.30 for video, and 0.40 / 0.50 for multimodal. For DaiKon Turn-Taking, the reported test performance is 0.66 Macro-F1 / 1.50 MAE for audio, 0.51 / 1.55 for video, and 0.63 / 1.50 for multimodal. For DaiKon Rapport, the reported test performance is 0.68 CCC / 0.70 Pearson for audio, 0.26 / 0.31 for video, and 0.59 / 0.64 for multimodal.

These baselines provide reference points for future submissions and indicate that audio carries strong information across all three tasks. At the same time, the challenge is structured to encourage further work on temporal modeling, dyad-aware architectures, and multimodal fusion strategies tailored to interpersonal interaction.

\section{Experiments}

\subsection{Experimental Setup}
All experiments follow the official train, validation, and test splits of
the Hume-DaiKon corpus ($661/142/142$ sessions). Models operate on the
standardized Parquet feature files distributed with the challenge:
768-dimensional Whisper-small~\cite{radford2023whisper} encoder embeddings
at approximately 3\,FPS from the combined audio, and 512-dimensional
FaceNet~\cite{schroff2015facenet} embeddings at 3\,FPS from per-speaker
face images, retaining only frames with detection probability $\geq 0.9$.
Multimodal baselines concatenate pooled audio with pooled per-speaker video.

All models are implemented in PyTorch~\cite{paszke2019pytorch} and
trained on a single NVIDIA H100 80\,GB GPU with a fixed random seed of 42.
Optimization uses AdamW with weight decay $10^{-4}$, dropout $0.2$, and
per-task learning rates selected on the validation split. Each baseline
uses a shared two-layer MLP encoder (Linear $\rightarrow$ ReLU $\rightarrow$
Dropout, repeated twice, with hidden dimension 256) followed by
task-specific output heads. The released baselines are intentionally kept
simple so that future work can focus on temporal, dyadic, and multimodal
extensions. Training uses early stopping with patience 5 based on the
validation metric. Model selection is performed on the validation split,
while the test set is reserved exclusively for official reporting.
Evaluation follows the task-specific protocol described in
Section~\ref{sec:tasks}: CCC and Pearson correlation for Influence and
Rapport, Macro-F1 and accuracy for next-speaker classification in
Turn-Taking, and MAE for time-to-next-speech regression.

\subsection{DaiKon Influence Baseline}
The influence baseline predicts target-speaker emotion intensity at the
VAD-segment level. For each segment $[t_s, t_e]$, audio and video
embeddings are mean-pooled over the segment. In the audio modality, the
pooled Whisper-small vector (768-D) is used directly. In the video
modality, the pooled FaceNet embeddings of the target speaker and the
partner speaker are concatenated (1024-D), providing dyadic visual
context. In the multimodal modality, pooled audio is concatenated with
both pooled per-speaker video features (1792-D). The joint representation
is passed through the shared MLP encoder and a final linear head with a
sigmoid nonlinearity, producing ten emotion intensities in $[0, 1]$. The
model is trained with mean squared error loss over all ten dimensions
jointly, with batch size 128, learning rate $5 \times 10^{-4}$, and up to
50 epochs. Validation metric for model selection: mean CCC across the ten
emotion dimensions.

\subsection{DaiKon Turn-Taking Baseline}
The turn-taking baseline operates at the event level: for each prediction
point in a conversation, features are mean-pooled over a fixed 10-second
context window ending at the prediction time. The pooled feature vector
is passed through the shared MLP encoder and two task heads: a linear
speaker head that performs binary classification (does the current
speaker continue, or does the floor transfer to the other speaker), and a
linear time head that regresses the time until the next speech onset,
clipped to $[-5, 10]$ seconds to accommodate both gaps and overlaps.
The model is trained with a joint objective combining binary
cross-entropy for the speaker head and L1 loss for the time head
(regression loss weight 0.1), with batch size 128, learning rate
$10^{-4}$, and up to 50 epochs.

\subsection{DaiKon Rapport Baseline}
The rapport baseline processes one labeled window per sample. Within each
window $[t_s, t_e]$, per-modality frame sequences are mean-pooled; for
video, both speakers' FaceNet embeddings are pooled and concatenated. The
resulting window descriptor (768-D for audio, 1024-D for video, 1792-D
for multimodal) is passed through the shared MLP encoder and a linear
head that produces a scalar rapport score. The model is trained with
mean squared error loss against the pseudo-label trajectories described
in Section~\ref{sec:rapport-labels}, with batch size 256, learning rate
$10^{-3}$, and up to 30 epochs. Validation metric for model selection: CCC.

\section{Results}
Table~\ref{tab:baseline_results} summarizes the baseline test results for all
three sub-challenges across modalities.


\begin{table*}[t]
\centering
\caption{Baseline test results across tasks and modalities.}
\label{tab:baseline_results}
\small
\setlength{\tabcolsep}{8pt}
\begin{tabular}{lrrrrrr}
\toprule
\textbf{Modality} & \multicolumn{2}{c}{\textbf{Influence}} & \multicolumn{2}{c}{\textbf{Rapport}} & \multicolumn{2}{c}{\textbf{Turn-Taking}} \\
\cmidrule{2-3} \cmidrule{4-5} \cmidrule{6-7}
 & \textbf{CCC} & \textbf{Pearson} & \textbf{CCC} & \textbf{Pearson} & \textbf{Macro-F1} & \textbf{MAE (s)} \\
\midrule
Audio      & 0.40 & 0.50 & 0.68 & 0.70 & 0.66 & 1.50 \\
Video      & 0.19 & 0.30 & 0.26 & 0.31 & 0.51 & 1.55 \\
Multimodal & 0.40 & 0.50 & 0.59 & 0.64 & 0.63 & 1.50 \\
\bottomrule
\end{tabular}
\end{table*}

\textbf{DaiKon Influence.} The audio and multimodal baselines achieve 0.40
CCC and 0.50 Pearson, substantially outperforming the video-only system
(0.19 CCC), suggesting that vocal cues are the primary carriers of
interpersonal influence signals in this dataset. The identical performance
of the audio and multimodal baselines indicates that, at this level of
pooling, video features do not add information beyond the audio signal alone.

\textbf{DaiKon Turn-Taking.} Audio gives the strongest results with 0.66
Macro-F1 and 1.50\,s MAE, consistent with the well-established role of
acoustic-prosodic cues in floor exchange. The multimodal system closely
matches the audio-only baseline, while the video system is the weakest.
The task is formulated as binary change detection — predicting whether the
current speaker continues or yields the floor — so a Macro-F1 of 0.66
should be interpreted against a random baseline near 0.50.

\textbf{DaiKon Rapport.} The audio-only baseline achieves 0.68 CCC and
0.70 Pearson. The video-only baseline is markedly weaker (0.26 CCC),
suggesting that mean-pooled FaceNet embeddings over a full rapport window
lose the fine-grained temporal structure needed to track interaction quality.
The multimodal baseline underperforms audio-only (0.59 CCC, 0.64 Pearson),
indicating that naive concatenation of pooled video adds noise rather than
complementary signal at this baseline level.

Across all three tasks, audio consistently outperforms video as a standalone
modality, and simple multimodal fusion does not consistently improve on the
audio-only baseline. These results highlight that richer temporal modeling,
dyad-aware architectures, and more sophisticated fusion strategies represent
the primary avenues for improvement.

\section{Conclusions}
The ACII-DaiKon Workshop \& Challenge establishes a benchmark for dyadic
interpersonal modeling in natural conversation. The three sub-challenges
jointly address directional affect, conversational timing, and rapport
trajectory modeling, encouraging methods that move beyond speaker-centric
prediction toward richer representations of inter-speaker dynamics. Baseline
results confirm that temporal modeling is essential across all tasks, that
audio carries the strongest signal in the current baselines, and that
multimodal fusion remains an open challenge. We invite cross-disciplinary
participation from researchers in affective computing, machine learning,
emotion science, conversation analysis, and cultural psychology.

\section*{Acknowledgment}
The challenge organizers would like to thank Hume AI for providing the dataset, as well as the Hume AI engineering team for their support in preparing and releasing the data. We also extend our gratitude to the participants whose contributions made this dataset possible.

\vspace{12pt}


\begin{thebibliography}{00}
\bibitem{zaki2013interpersonal} J. Zaki and W. C. Williams, ``Interpersonal emotion regulation,'' \emph{Emotion}, vol. 13, no. 5, pp. 803--810, 2013.
\bibitem{sacks1974turntaking} H. Sacks, E. A. Schegloff, and G. Jefferson, ``A simplest systematics for the organization of turn-taking for conversation,'' \emph{Language}, vol. 50, no. 4, pp. 696--735, 1974.
\bibitem{tickledegnen1990rapport} L. Tickle-Degnen and R. Rosenthal, ``The nature of rapport and its nonverbal correlates,'' \emph{Psychological Inquiry}, vol. 1, no. 4, pp. 285--293, 1990.
\bibitem{baltrusaitis2019multimodal} T. Baltru\v{s}aitis, C. Ahuja, and L.-P. Morency, ``Multimodal machine learning: A survey and taxonomy,'' \emph{IEEE Trans. Pattern Anal. Mach. Intell.}, vol. 41, no. 2, pp. 423--443, 2019.
\bibitem{picard1997affective} R. W. Picard, \emph{Affective Computing}. Cambridge, MA, USA: MIT Press, 1997.
\bibitem{radford2023whisper} A. Radford \emph{et al.}, ``Robust speech recognition via large-scale weak supervision,'' in \emph{Proc. ICML}, 2023.
\bibitem{schroff2015facenet} F. Schroff, D. Kalenichenko, and J. Philbin, ``FaceNet: A unified embedding for face recognition and clustering,'' in \emph{Proc. CVPR}, 2015.
\bibitem{qwen2025}
Qwen Team, ``Qwen2.5 Technical Report,'' arXiv:2412.15115, 2025.
\bibitem{kwon2023vllm}
W.~Kwon \emph{et al.}, ``Efficient memory management for large language model
serving with PagedAttention,'' in \emph{Proc. ACM SOSP}, 2023.
\bibitem{javed2024dyadic} H. Javed, W. Wang, A. B. Usman, and N. Jamali, ``Modeling interpersonal perception in dyadic interactions: towards robot-assisted social mediation in the real world,'' \emph{Frontiers in Robotics and AI}, vol. 11, p. 1410957, 2024.
\bibitem{song2025affective} X. Song, C. Liu, L. Xu, Z. Lu, and Y. Zhang, ``Affective computing methods for multimodal embodied AI human--computer interaction,'' \emph{Aslib Journal of Information Management}, 2025.
\bibitem{paszke2019pytorch} A. Paszke \emph{et al.}, ``PyTorch: An imperative style, high-performance deep learning library,'' in \emph{Advances in Neural Information Processing Systems 32}, 2019.
\bibitem{b1} G. Eason, B. Noble, and I. N. Sneddon, ``On certain integrals of Lipschitz-Hankel type involving products of Bessel functions,'' Phil. Trans. Roy. Soc. London, vol. A247, pp. 529--551, April 1955.
\bibitem{b2} J. Clerk Maxwell, A Treatise on Electricity and Magnetism, 3rd ed., vol. 2. Oxford: Clarendon, 1892, pp.68--73.
\bibitem{b3} I. S. Jacobs and C. P. Bean, ``Fine particles, thin films and exchange anisotropy,'' in Magnetism, vol. III, G. T. Rado and H. Suhl, Eds. New York: Academic, 1963, pp. 271--350.
\bibitem{b4} K. Elissa, ``Title of paper if known,'' unpublished.
\bibitem{b5} R. Nicole, ``Title of paper with only first word capitalized,'' J. Name Stand. Abbrev., in press.
\bibitem{b6} Y. Yorozu, M. Hirano, K. Oka, and Y. Tagawa, ``Electron spectroscopy studies on magneto-optical media and plastic substrate interface,'' IEEE Transl. J. Magn. Japan, vol. 2, pp. 740--741, August 1987 [Digests 9th Annual Conf. Magnetics Japan, p. 301, 1982].
\bibitem{b7} M. Young, The Technical Writer's Handbook. Mill Valley, CA: University Science, 1989.

\bibitem{russell1980circumplex}
J.~A. Russell, ``A circumplex model of affect,''
\emph{Journal of Personality and Social Psychology}, vol. 39, no. 6,
pp. 1161--1178, 1980.

\bibitem{baird2022icml} A. Baird, P. Tzirakis, G. Gidel, M. Jiralerspong, E. B. Muller, K. Mathewson, B. Schuller, E. Cambria, D. Keltner, and A. Cowen, ``The ICML 2022 Expressive Vocalizations Workshop and Competition: Recognizing, generating, and personalizing vocal bursts,'' in \emph{Proc. ICML Expressive Vocalizations Workshop and Competition}, 2022.

\bibitem{tzirakis2023largescale} P. Tzirakis, A. Baird, J. Brooks, C. Gagne, L. Kim, M. Opara, C. Gregory, J. Metrick, G. Boseck, V. Tiruvadi, B. Schuller, D. Keltner, and A. Cowen, ``Large-scale nonverbal vocalization detection using transformers,'' in \emph{ICASSP 2023 -- IEEE Int. Conf. Acoustics, Speech and Signal Processing}, pp. 1--5, 2023.

\bibitem{stappen2020muse} L. Stappen, A. Baird, G. Rizos, P. Tzirakis, X. Du, F. Hafner, L. Schumann, A. Mallol-Ragolta, B. W. Schuller, I. Lefter, E. Cambria, and I. Kompatsiaris, ``MuSe 2020 Challenge and Workshop: Multimodal sentiment analysis, emotion-target engagement and trustworthiness detection in real-life media,'' in \emph{Proc. 1st Int. Multimodal Sentiment Analysis in Real-life Media Challenge and Workshop}, pp. 35--44, 2020.

\bibitem{christ2023muse} L. Christ, S. Amiriparian, A. Baird, A. Kathan, N. M\"uller, S. Klug, C. Gagne, P. Tzirakis, L. Stappen, E.-M. Me\ss{}ner, A. K\"onig, A. Cowen, E. Cambria, and B. W. Schuller, ``The MuSe 2023 Multimodal Sentiment Analysis Challenge: Mimicked emotions, cross-cultural humour, and personalisation,'' in \emph{Proc. 4th Multimodal Sentiment Analysis Challenge and Workshop}, 2023.

\bibitem{tzirakis2017end} P. Tzirakis, G. Trigeorgis, M. A. Nicolaou, B. W. Schuller, and S. Zafeiriou, ``End-to-end multimodal emotion recognition using deep neural networks,'' \emph{IEEE Journal of Selected Topics in Signal Processing}, vol. 11, no. 8, pp. 1301--1309, 2017.
 
\end{thebibliography}
\end{document}